# Compiling HPSG type constraints into definite clause programs

Thilo Götz and Walt Detmar Meurers*
SFB 340, Universität Tübingen
Kleine Wilhelmstraße 113
72074 Tübingen
Germany
{tg,dm}@sfs.nphil.uni-tuebingen.de

## Abstract

We present a new approach to HPSG processing: compiling HPSG grammars expressed as type constraints into definite clause programs. This provides a clear and computationally useful correspondence between linguistic theories and their implementation. The compiler performs off-line constraint inheritance and code optimization. As a result, we are able to efficiently process with HPSG grammars without having to hand-translate them into definite clause or phrase structure based systems.

## 1 Introduction

The HPSG architecture as defined in (Pollard and Sag, 1994) (henceforth HPSG II) is being used by an increasing number of linguists, since the formally well-defined framework allows for a rigid and explicit formalization of a linguistic theory. At the same time, the feature logics which provide the formal foundation of HPSG II have been used as basis for several NLP systems, such as ALE (Carpenter, 1993), CUF (Dörre and Dorna, 1993), Troll (Gerdemann and King, 1993) or TFS (Emele and Zajac, 1990). These systems are – at least partly – intended as computational environments for the implementation of HPSG grammars.

HPSG linguists use the description language of the logic to express their theories in the form of implicative constraints. On the other hand, most of the computational setups only allow feature descriptions as extra constraints with a phrase structure or definite clause based language.[1] From a computational point of view the latter setup has several advantages. It provides access to the pool of work done in the area of natural language processing, e.g., to efficient control strategies for the definite clause level based on tabelling methods like Earley deduction, or different parsing strategies in the phrase structure setup.

The result is a gap between the description language theories of HPSG linguists and the definite clause or phrase structure based NLP systems provided to implement these theories. Most grammars currently implemented therefore have no clear correspondence to the linguistic theories they originated from. To be able to use implemented grammars to provide feedback for a rigid and complete formalization of linguistic theories, a clear and computationally useful correspondence has to be established. This link is also needed to stimulate further development of the computational systems. Finally, an HPSG II style setup is also interesting to model from a software engineering point of view, since it permits a modular development and testing of the grammar.

The purpose of this paper is to provide the desired link, i.e., to show how a HPSG theory formulated as implicative constraints can be modelled on the level of the relational extension of the constraint language. More specifically, we define a compilation procedure which translates the type constraints of the linguistic theory into definite clauses runnable in systems such as Troll, ALE, or CUF. Thus, we perform constraint inheritance and code optimization off-line. This results in a considerable efficiency gain over a direct on-line treatment of type constraints as, e.g., in TFS.

The structure of the paper is as follows: A short discussion of the logical setup for HPSG II provides the necessary formal background and terminology. Then the two possibilities for expressing a theory – using the description language as in HPSG II or the relational level as in the computational architectures – are introduced. The third section provides a simple picture of how HPSG II theories can be modelled on the relational level. This simple picture is then refined in the fourth section, where the compilation procedure and its implementation is discussed. A small example grammar is provided in the appendix.

---

*The authors are listed alphabetically.

[1] One exception is the TFS system. However, the possibility to express recursive relations on the level of the description language leads to serious control problems in that system.

## 2 Background

### 2.1 The HPSG II architecture

A HPSG grammar consists of two components: the declaration of the structure of the domain of linguistic objects in a signature (consisting of the type hierarchy and the appropriateness conditions) and the formulation of constraints on that domain. The signature introduces the structures the linguist wants to talk about. The theory the linguist proposes distinguishes between those objects in a domain which are part of the natural language described, and those which are not.

HPSG II gives a closed world interpretation to the type hierarchy: every object is of exactly one minimal (most specific) type. This implies that every object in the denotation of a non-minimal type is also described by at least one of its subtypes. Our compilation procedure will adhere to this interpretation.

### 2.2 The theories of HPSG II: Directly constraining the domain

A HPSG II theory consists of a set of descriptions which are interpreted as being true or false of an object in the domain. An object is admissible with respect to a certain theory iff it satisfies each of the descriptions in the theory and so does each of its substructures. The descriptions which make up the theory are also called *constraints*, since these descriptions constrain the set of objects which are admissible with respect to the theory.

Figure 1 shows an example of a constraint, the head-feature principle of HPSG II. Throughout the paper we will be using HPSG style AVM notation for descriptions.

$$\begin{bmatrix} phrase \\ \text{DTRS} \ headed\text{-}struc \end{bmatrix} \rightarrow$$

$$\begin{bmatrix} \text{SYNSEM|LOC|CAT|HEAD} & \boxed{1} \\ \text{DTRS|HEAD-DTR|SYNSEM|LOC|CAT|HEAD} & \boxed{1} \end{bmatrix}$$

Figure 1: The Head-Feature Principle of HPSG II

The intended interpretation of this constraint is that every object which is being described by type *phrase* and by [DTRS *headed-struc*] also has to be described by the consequent, i.e. have its head value shared with that of its head-daughter.

In the HPSG II architecture any description can be used as antecedent of an implicative constraint. As shown in (Meurers, 1994), a complex description can be expressed as a type by modifying the signature and/or adding theory statements. In the following, we therefore only deal with implicative constraints with type antecedents, the type definitions.

### 2.3 Theories in constraint logic programming: expressing definite clause relations

As mentioned in the introduction, in most computational systems for the implementation of HPSG theories a grammar is expressed using a relational extension of the description language[2] such as definite clauses or phrase structure rules. Figure 2 schematically shows the embedding of HPSG II descriptions in the definition of a relation.

$$rel_0(D_1, \ldots, D_i) \ :- \ rel_1(E_1, \ldots, E_j),$$
$$\vdots$$
$$rel_n(F_1, \ldots, F_k).$$

Figure 2: Defining relation $rel_0$

The HPSG description language is only used to specify the arguments of the relations, in the example noted as $D$, $E$, and $F$. The organization of the descriptions, i.e. their use as constraints to narrow down the set of described objects, is taken over by the relational level. This way of organizing descriptions in definite clauses allows efficient processing techniques of logic programming to be used.

The question we are concerned with in the following is how a HPSG II theory can be modelled in such a setup.

## 3 Modelling HPSG II theories on a relational level: a simple picture

There are three characteristics of HPSG II theories which we need to model on the relational level: one needs to be able to

1. express constraints on any kind of object,

2. use the hierarchical structure of the type hierarchy to organize the constraints, and

3. check any structure for consistency with the theory.

A straightforward encoding is achieved by expressing each of these three aspects in a set of relations. Let us illustrate this idea with a simple example. Assume the signature given in figure 3 and the HPSG II

---

[2] For the logical foundations of relational extensions of arbitrary constraint languages see (Höhfeld and Smolka, 1988).

style theory of figure 4.

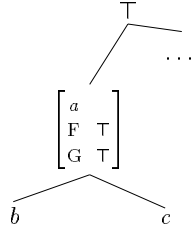

Figure 3: An example signature

$$a \rightarrow \begin{bmatrix} F & \boxed{1} \\ G & \boxed{1} \end{bmatrix}$$

$$b \rightarrow \begin{bmatrix} G & a \end{bmatrix}$$

Figure 4: An example theory in a HPSG II setup

First, we define a relation to express the constraints immediately specified for a type on the argument of the relation:

- $a_{cons}(\begin{bmatrix} a \\ F & \boxed{1} \\ G & \boxed{1} \end{bmatrix})$ :- $\top_{type}(\boxed{1})$.

- $b_{cons}(\begin{bmatrix} b \\ G & \boxed{1} \end{bmatrix})$ :- $a_{type}(\boxed{1})$.

- $c_{cons}(c)$.

For every type, the relation specifies its only argument to bear the type information and the consequents of the type definition for that type. Note that the simple type assignment [G $a$] leads to a call to the relation $a_{type}$ imposing all constraints for type $a$, which is defined below.

Second, a relation is needed to capture the hierarchical organization of constraints:

- $\top_{hier}(\boxed{1})$ :- $a_{hier}(\boxed{1})$ ; ....[3]

- $a_{hier}(\boxed{1})$ :- $a_{cons}(\boxed{1})$, ( $b_{hier}(\boxed{1})$; $c_{hier}(\boxed{1})$ ).

- $b_{hier}(\boxed{1})$ :- $b_{cons}(\boxed{1})$.

- $c_{hier}(\boxed{1})$ :- $c_{cons}(\boxed{1})$.

Each hierarchy relation of a type references the constraint relation and makes sure that the constraints below one of the subtypes are obeyed.

Finally, a relation is defined to collect all constraints on a type:

- $a_{type}(\boxed{1})$ :- $\top_{hier}(\boxed{1}a)$.

- $b_{type}(\boxed{1})$ :- $\top_{hier}(\boxed{1}b)$.

- $c_{type}(\boxed{1})$ :- $\top_{hier}(\boxed{1}c)$.

---

[3] A disjunction of the immediate subtypes of $\top$.

Compared to the hierarchy relation of a type which collects all constraints on the type and its subtypes, the last kind of relation additionally references those constraints which are inherited from a supertype. Thus, this is the relation that needs to be queried to check for grammaticality.

Even though the simple picture with its tripartite definition for each type yields perspicuous code, it falls short in several respects. The last two kinds of relations (rel$_{type}$ and rel$_{hier}$) just perform inheritance of constraints. Doing this at run-time is slow, and additionally there are problems with multiple inheritance.

A further problem of the encoding is that the value of an appropriate feature which is not mentioned in any type definition may nonetheless be implicitly constrained, since the type of its value is constrained. Consider for example the standard HPSG encoding of list structures. This usually involves a type *ne_list* with appropriate features HD and TL, where under HD we encode an element of the list, and under TL the tail of the list. Normally, there will be no extra constraints on *ne_list*. But in our setup we clearly need a definite clause

$$\text{ne\_list}_{cons}(\begin{bmatrix} ne\_list \\ HD & \boxed{1} \\ TL & \boxed{2} \end{bmatrix})$$ :- $\top_{type}(\boxed{1})$, $\text{list}_{type}(\boxed{2})$.

since the value of the feature HD may be of a type which is constrained by the grammar. Consequently, since *ne_list* is a subtype of *list*, the value of TL needs to be constrained as well.

## 4 Compiling HPSG type constraints into definite clauses

After this intuitive introduction to the problem, we will now show how to automatically generate definite clause programs from a set of type definitions, in a way that avoids the problems mentioned for the simple picture.

### 4.1 The algorithm

Before we can look at the actual compilation procedure, we need some terminology.

**Definition** (type interaction)

*Two types interact if they have a common subtype.*

Note that every type interacts with itself.

**Definition** (defined type)

*A defined type is a type that occurs as antecedent of an implicational constraint in the grammar.*

**Definition** (constrained type)

*A constrained type is a type that interacts with a defined type.*

Whenever we encounter a structure of a constrained type, we need to check that the structure conforms to the constraint on that type. As mentioned in section 2.1, due to the closed world interpretation of type hierarchies, we know that every object in the denotation of a non-minimal type $t$ also has to obey the constraints on one of the minimal subtypes of $t$. Thus, if a type $t$ has a subtype $t'$ in common with a defined type $d$, then $t'$ is a constrained type (by virtue of being a subtype of $d$) and $t$ is a constrained type (because it subsumes $t'$).

**Definition** (hiding type)

*The set of hiding types is the smallest set s.t. if $t$ is not a constrained type and subsumes a type $t_0$ that has a feature $f$ appropriate s.t. approp($t_0$,$f$) is a constrained type or a hiding type, then $t$ is a hiding type.*

The type *ne_list* that we saw above is a hiding type.

**Definition** (hiding feature)

*If $t$ is a constrained or hiding type, then $f$ is a hiding feature on $t$ iff approp($t$,$f$) is a constrained or hiding type.*

**Definition** (simple type)

*A simple type is a type that is neither a constrained nor a hiding type.*

When we see a structure of a simple type, we don't need to apply any constraints, neither on the top node nor on any substructure.

Partitioning the types in this manner helps us to construct definite clause programs for type constraint grammars. For each type, we compute a unary relation that we just give the same name as the type. Since we assume a closed world interpretation of the type hierarchy, we really only need to compute proper definitions for minimal types. The body of a definition for a non-minimal type is just a disjunction of the relations defining the minimal subtypes of the non-minimal type.

When we want to compute the defining clause for a minimal type, we first of all check what sort of type it is. For each simple type, we just introduce a unit clause whose argument is just the type. For a constrained type $t$, first of all we have to perform constraint inheritance from all types that subsume $t$. Then we transform that constraint to some internal representation, usually a feature structure (FS). We now have a schematic defining clause of the form

$$t(FS) :- ?.$$

Next, we compute the missing right-hand side (RHS) with the following algorithm.

1. Compute HF, the set of hiding features on the type of the current node, then insert these features with appropriate types in the structure (FS) if they're not already there. For each node under a feature in HF, apply step 2.
2. Let $t$ be the type on the current node and X its tag (a variable).
   (a) If $t$ is a constrained type, enter $t(X)$ into RHS (if it's not already there).
   (b) Elseif $t$ is a hiding type, then check if its hiding features and the hiding features of all its hiding subtypes are identical. If they are identical, then proceed as in step 1. If not, enter $t(X)$ into RHS.
   (c) Else ($t$ is a simple type) do nothing at all.

For hiding types, we do exactly the same thing, except that we don't have any structure to begin with. But this is no problem, since the hiding features get introduced anyway.

### 4.2 An example

A formal proof of correctness of this compiler is given in (Götz, 1995) – here, we will try to show by example how it works. Our example is an encoding of a definite relation in a type constraint setup.[4] *append_c* appends an arbitrary list onto a list of constants.

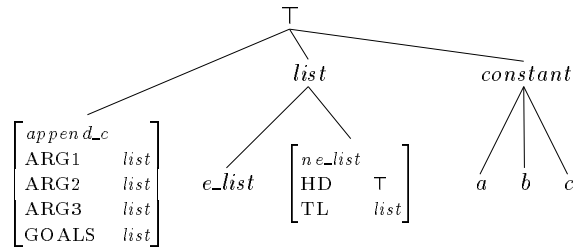

Figure 5: The signature for the *append_c* example

We will stick to an AVM style notation for our examples, the actual program uses a standard feature term syntax. List are abbreviated in the standard HPSG manner, using angled brackets.

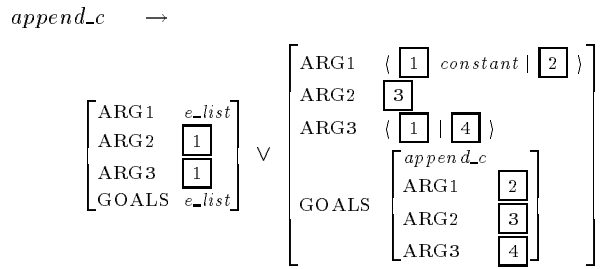

Figure 6: A constraint on *append_c*

Note that the set of constrained types is {*append_c*,

---

[4] This sort of encoding was pioneered by (Aït-Kaci, 1984), but see also (King, 1989) and (Carpenter, 1992).

⊤} and the set of hiding types is {*list*, *ne_list*}. Converting the first disjunct of *append_c* into a feature structure to start our compilation, we get something like

$$\text{append\_c}\left(\begin{bmatrix} append\_c \\ \text{ARG1} & \boxed{1} & e\_list \\ \text{ARG2} & \boxed{2} & list \\ \text{ARG3} & \boxed{2} & \\ \text{GOALS} & \boxed{3} & e\_list \end{bmatrix}\right) \text{ :- ?.}$$

Since the values of the features of *append_c* are of type *list*, a hiding type, those features are hiding features and need to be considered. Yet looking at node $\boxed{1}$, the algorithm finds *e_list*, a simple type, and does nothing. Similarly with node $\boxed{3}$. On node $\boxed{2}$, we find the hiding type *list*. Its one hiding subtype, *ne_list*, has different hiding features (*list* has no features appropriate at all). Therefore, we have to enter this node into the RHS. Since the same node appears under both ARG1 and ARG2, we're done and have

$$\text{append\_c}\left(\begin{bmatrix} append\_c \\ \text{ARG1} & & e\_list \\ \text{ARG2} & \boxed{2} & list \\ \text{ARG3} & \boxed{2} & \\ \text{GOALS} & & e\_list \end{bmatrix}\right) \text{ :- list}(\boxed{2}).$$

which is exactly what we want. It means that a structure of type *append_c* is well-formed if it unifies with the argument of the head of the above clause and whatever is under ARG2 (and ARG3) is a well-formed list. Now for the recursive disjunct, we start out with

$$\text{append\_c}\left(\begin{bmatrix} append\_c \\ \text{ARG1} & \boxed{1} & \begin{bmatrix} ne\_list \\ \text{HD} & \boxed{2} & constant \\ \text{TL} & \boxed{3} & list \end{bmatrix} \\ \text{ARG2} & \boxed{4} & list \\ \text{ARG3} & \boxed{5} & \begin{bmatrix} ne\_list \\ \text{HD} & \boxed{2} & \\ \text{TL} & \boxed{6} & list \end{bmatrix} \\ \text{GOALS} & \boxed{7} & \begin{bmatrix} ne\_list \\ \text{HD} & \boxed{8} & \begin{bmatrix} append\_c \\ \text{ARG1} & \boxed{3} \\ \text{ARG2} & \boxed{4} \\ \text{ARG3} & \boxed{6} \end{bmatrix} \\ \text{TL} & \boxed{9} & e\_list \end{bmatrix} \end{bmatrix}\right) \text{ :- ?.}$$

Node $\boxed{1}$ bears a hiding type with no subtypes. Therefore we don't enter that node in the RHS, but proceed to look at its features. Node $\boxed{2}$ bears a simple type and we do nothing, but node $\boxed{3}$ is again a *list* and needs to be entered into the RHS. Similarly with nodes $\boxed{4}$ and $\boxed{6}$. *append_c* on node $\boxed{8}$ is a constrained type and $\boxed{8}$ also has to go onto the RHS.

The final result then is

$$\text{append\_c}\left(\begin{bmatrix} append\_c \\ \text{ARG1} & \begin{bmatrix} ne\_list \\ \text{HD} & \boxed{2} & constant \\ \text{TL} & \boxed{3} & list \end{bmatrix} \\ \text{ARG2} & \boxed{4} & list \\ \text{ARG3} & \begin{bmatrix} ne\_list \\ \text{HD} & \boxed{2} \\ \text{TL} & \boxed{6} & list \end{bmatrix} \\ \text{GOALS} & \begin{bmatrix} ne\_list \\ \text{HD} & \boxed{8} & \begin{bmatrix} append\_c \\ \text{ARG1} & \boxed{3} \\ \text{ARG2} & \boxed{4} \\ \text{ARG3} & \boxed{6} \end{bmatrix} \\ \text{TL} & e\_list \end{bmatrix} \end{bmatrix}\right) \text{ :-}$$

list($\boxed{3}$), list($\boxed{4}$), list($\boxed{6}$), append_c($\boxed{8}$).

This is almost what we want, but not quite. Consider node $\boxed{8}$. Clearly it needs to be checked, but what about nodes $\boxed{3}$, $\boxed{4}$ and $\boxed{6}$? They are all embedded under node $\boxed{8}$ which is being checked anyway, so listing them here in the RHS is entirely redundant. In general, if a node is listed in the RHS, then no other node below it needs to be there as well. Thus, our result should really be

$$\text{append\_c}\left(\begin{bmatrix} append\_c \\ \text{ARG1} & \begin{bmatrix} ne\_list \\ \text{HD} & \boxed{2} & constant \\ \text{TL} & \boxed{3} & list \end{bmatrix} \\ \text{ARG2} & \boxed{4} & list \\ \text{ARG3} & \begin{bmatrix} ne\_list \\ \text{HD} & \boxed{2} \\ \text{TL} & \boxed{6} & list \end{bmatrix} \\ \text{GOALS} & \begin{bmatrix} ne\_list \\ \text{HD} & \boxed{8} & \begin{bmatrix} append\_c \\ \text{ARG1} & \boxed{3} \\ \text{ARG2} & \boxed{4} \\ \text{ARG3} & \boxed{6} \end{bmatrix} \\ \text{TL} & e\_list \end{bmatrix} \end{bmatrix}\right) \text{ :-}$$

append_c($\boxed{8}$).

Our implementation of the compiler does in fact perform this pruning as an integrated part of the compilation, not as an additional step.

It should be pointed out that this compilation result is quite a dramatic improvement on more naive on-line approaches to HPSG processing. By reasoning with the different kinds of types, we can drastically reduce the number of goals that need to be checked on-line. Another way of viewing this would be to see the actual compilation step as being much simpler (just check every possible feature) and to subsequently apply program transformation techniques (some sophisticated form of partial evaluation). We believe that this view would not simplify the overall picture, however.

### 4.3 Implementation and Extensions

The compiler as described in the last section has been fully implemented under Quintus Prolog. Our interpreter at the moment is a simple left to right backtracking interpreter. The only extension is to keep a list of all the nodes that have already been visited to keep the same computation from being repeated. This is necessary since although we avoid redundancies as shown in the last example, there are still cases where the same node gets checked more than once.

This simple extension also allows us to process cyclic queries. The following query is allowed by our system.

$$\text{Query>} \quad \boxed{1} \begin{bmatrix} ne\_list \\ \text{HD} & a \\ \text{TL} & \boxed{1} \end{bmatrix}$$

Figure 7: A permitted cyclic query

An interpreter without the above-mentioned extension would not terminate on this query.

The computationally oriented reader will now wonder how we expect to deal with non-termination anyway. At the moment, we allow the user to specify minimal control information.

- The user can specify an ordering on type expansion. E.g., if the type hierarchy contains a type *sign* with subtypes *word* and *phrase*, the user may specify that *word* should always be tried before *phrase*.

- The user can specify an ordering on feature expansion. E.g., HD should always be expanded before TL in a given structure.

Since this information is local to any given structure, the interpreter does not need to know about it, and the control information is interpreted as compiler directives.

## 5 Conclusion and Outlook

We have presented a compiler that can encode HPSG type definitions as a definite clause program. This for the first time offers the possibility to express linguistic theories the way they are formulated by linguists in a number of already existing computational systems.

The compiler finds out exactly which nodes of a structure have to be examined and which don't. In doing this off-line, we minimize the need for on-line inferences. The same is true for the control information, which is also dealt with off-line. This is not to say that the interpreter wouldn't profit by a more sophisticated selection function or tabulation techniques (see, e.g., (Dörre, 1993)). We plan to apply Earley deduction to our scheme in the near future and experiment with program transformation techniques and bottom-up interpretation.

Our work addresses a similar problem as Carpenter's work on resolved feature structures (Carpenter, 1992, ch. 15). However, there are two major differences, both deriving form the fact that Carpenter uses an open world interpretation. Firstly, our approach can be extended to handle arbitrarily complex antecedents of implications (i.e., arbitrary negation), which is not possible using an open world approach. Secondly, solutions in our approach have the so-called *subsumption monotonicity* or *persistence* property. That means that any structure subsumed by a solution is also a solution (as in Prolog, for example). Quite the opposite is the case in Carpenter's approach, where solutions are not guaranteed to have more specific extensions. This is unsatisfactory at least from an HPSG point of view, since HPSG feature structures are supposed to be maximally specific.

## Acknowledgments

The research reported here was carried out in the context of SFB 340, project B4, funded by the Deutsche Forschungsgemeinschaft. We would like to thank Dale Gerdemann, Paul John King and two anonymous referees for helpful discussion and comments.

## Appendix A. A small grammar

The following small example grammar, together with a definition of an append type, generates sentences like "John thinks cats run". It is a modified version of an example from (Carpenter, 1992).

$$
phrase \rightarrow
$$

$$
\left( \begin{bmatrix} \text{CAT} & s \\ \text{DTR1} & \begin{bmatrix} \text{CAT} & np \\ \text{AGR} & \boxed{1} \\ \text{PHON} & \boxed{2} \end{bmatrix} \\ \text{DTR2} & \begin{bmatrix} \text{CAT} & vp \\ \text{AGR} & \boxed{1} \\ \text{PHON} & \boxed{3} \end{bmatrix} \end{bmatrix} \vee \begin{bmatrix} \text{CAT} & vp \\ \text{AGR} & \boxed{1} \\ \text{DTR1} & \begin{bmatrix} \text{CAT} & sv \\ \text{AGR} & \boxed{1} \\ \text{PHON} & \boxed{2} \end{bmatrix} \\ \text{DTR2} & \begin{bmatrix} \text{CAT} & s \\ \text{PHON} & \boxed{3} \end{bmatrix} \end{bmatrix} \right)
$$

$$
\wedge \begin{bmatrix} \text{PHON} & \boxed{4} \\ \text{GOALS} & \langle \begin{bmatrix} append \\ \text{ARG1} & \boxed{2} \\ \text{ARG2} & \boxed{3} \\ \text{ARG3} & \boxed{4} \end{bmatrix} \rangle \end{bmatrix}
$$

$$
word \rightarrow \begin{bmatrix} \text{CAT} & np \\ \text{PHON} & \langle john \vee mary \rangle \\ \text{AGR} & singular \end{bmatrix}
$$

$$
\vee \begin{bmatrix} \text{CAT} & np \\ \text{PHON} & \langle cats \vee dogs \rangle \\ \text{AGR} & plural \end{bmatrix}
$$

$$
\vee \begin{bmatrix} \text{CAT} & vp \\ \text{PHON} & \langle runs \vee jumps \rangle \\ \text{AGR} & singular \end{bmatrix}
$$

$$
\vee \begin{bmatrix} \text{CAT} & vp \\ \text{PHON} & \langle run \vee jump \rangle \\ \text{AGR} & plural \end{bmatrix}
$$

$$
\vee \begin{bmatrix} \text{CAT} & sv \\ \text{PHON} & \langle knows \vee thinks \rangle \\ \text{AGR} & singular \end{bmatrix}
$$

Here's an example query. Note that the feature GOALS has been suppressed in the result.

**Query>** $\begin{bmatrix} \text{PHON} & \langle john, runs \rangle \end{bmatrix}$

**Result>** $\begin{bmatrix} phrase \\ \text{CAT} & s \\ \text{PHON} & \langle \boxed{1} john \mid \boxed{2} \langle runs \rangle \rangle \\ \text{DTR1} & \begin{bmatrix} word \\ \text{CAT} & np \\ \text{AGR} & \boxed{3} singular \\ \text{PHON} & \langle \boxed{1} \rangle \end{bmatrix} \\ \text{DTR2} & \begin{bmatrix} word \\ \text{CAT} & vp \\ \text{AGR} & \boxed{3} \\ \text{PHON} & \boxed{2} \end{bmatrix} \end{bmatrix}$

For the next query we get exactly the same result.

**Query>** $\begin{bmatrix} \text{DTR1} & \begin{bmatrix} \text{PHON} & \langle john \rangle \end{bmatrix} \\ \text{DTR2} & \begin{bmatrix} \text{PHON} & \langle runs \rangle \end{bmatrix} \end{bmatrix}$